# REGULARIZED FUZZY NEURAL NETWORKS TO AID EFFORT FORECASTING IN THE CONSTRUCTION AND SOFTWARE DEVELOPMENT

Paulo Vitor de Campos Souza, Augusto Junio Guimaraes, Vanessa Souza Araujo, Thiago Silva Rezende, Vinicius Jonathan Silva Araujo

Avenue Amazonas 5253, Belo Horizonte, Brazil
CEFET-MG [1,1,1], *Faculty Una Betim_
[a]Avenue. Governador Valadares, 640 - Centro,
[b]Betim - MG, 32510-010.

## ABSTRACT

*Predicting the time to build software is a very complex task for software engineering managers. There are complex factors that can directly interfere with the productivity of the development team. Factors directly related to the complexity of the system to be developed drastically change the time necessary for the completion of the works with the software factories. This work proposes the use of a hybrid system based on artificial neural networks and fuzzy systems to assist in the construction of an expert system based on rules to support in the prediction of hours destined to the development of software according to the complexity of the elements present in the same. The set of fuzzy rules obtained by the system helps the management and control of software development by providing a base of interpretable estimates based on fuzzy rules. The model was submitted to tests on a real database, and its results were promissory in the construction of an aid mechanism in the predictability of the software construction.*

## KEYWORDS

*Fuzzy neural networks, effort forecasting, use case point, expert systems,*

## 1. INTRODUCTION

Software development tracks the growth of technology worldwide. The continuing need for systems that facilitate human life grows as the world population has more access to electronic resources. The construction of systems is an area 5 that has been increasing its participation in the market, mainly with the awareness by companies and the government that the operational efficiency must be present in several areas of its performance [1]. However, managing the construction of software is not a simple task. Factors can disrupt the construction of software efficiently such as team expertise, the complexity of use cases used 10 or even the lack of standardization of work by developers. Numerous factors undermine the time estimates made by software managers, hampering financial and productivity estimates [2]. A study has been carried out to collect data referring to several multiple criteria present in software development. These studies have begun evaluations to understand the nature of this problem and to 15 build

---

mechanisms that allow managers to be able to close the deadline for delivering a software product to their client [3]. This paper proposes to use a real database collected in Turkish software development companies [4] for a hybrid model of fuzzy neural networks, to construct a system of rules to assist in the construction of expert systems for the resolution of problems of time estimation 20 of software development [5]. The fuzzy rules are constructed through the first layer of the hybrid model that uses concepts from the Anfis algorithm [6] to create equally spaced membership functions for each input variable. This approach allows creating Takagi Sugeno rules of the IF / THEN type [7], allowing the database to be interpreted more straightforwardly by the managers involved in the process. The second layer uses fuzzy logic neurons, which perform the aggregation of the neurons of the  of the first layer. Because it is a problem where the number of neurons generated can be high due to the number of characteristics evaluated, a bootstrap-based smoothing technique is applied to define the essential neurons of the problem. The final weights of the second layer will be used by a neural aggregation network that has a single artificial neuron [5].The synaptic weights will be generated using the concept of extreme learning machine [8], where there is no need to update the internal parameters of the fuzzy neural network continually. The paper is organized as follows: Section II presents the central concepts that guide the research, such as the definitions of fuzzy neural networks, concepts related to the complexity of software construction. Section III will have as its central focus the presentation of aspects related to the database and how it has already been worked in the literature, evidencing the contribution of this work. In it will also be present concepts and forms with which the model of fuzzy neural networks will act in the resolution of the problems. Section IV will present the methodology used in the tests and the methodologies, parameters, and configurations of the tests and their respective results will be presented. Finally, in section V the conclusions of the work will be presented.

## 2. LITERATURE REVIEW

### 2.1. Software development

Software development has significantly evolved through the introduction of new technologies. The use of mobile phones, direct access to information and agility in business requires that this production is increasingly dynamic and cohesive to meet the emerging needs of this target audience. Nowadays, software production is an area of study of several professors and researchers in the area of software engineering, allowing new techniques of inspection, evaluation, and improvement of activities in the construction of applications are a current aspect in software factories [1]. However, there is a specific barrier to efficiency in software production and customer satisfaction. In surveys carried out by [9] and [10] it is noted that the majority of customers are dissatisfied with the services provided by software factories, mainly stand out the complaints due to lack of organization, product out of compliance and main delays in deliveries.

Software development suffers from immeasurable external variables such as the lack of technological resources to develop a specific task requested by the contractor, lack of skilled labor, lack of tools in one language, high volatility of customer requests during the phase of project execution, among others. The complexity of the system itself is a factor that cannot be disregarded. The time for the production of a use case varies according to several external agents, but the complexity of it interferes directly in this type of evaluation [11]. As Well, as the use cases, the complexity of the actors involved in the problem, the connection between internal and external systems, the use of techniques such as refactoring and reusing are highlighted. It is not a fact that several techniques were developed to favor agile development over traditional development. These alternatives can improve delivery time, but your forecasting still becomes a complex factor, depending solely on the experience of the project manager, who must know very well the potential of his team and the level of complexity of the system that is responsible [2].

## 2.2. Use Case Points

Several techniques have been proposed to facilitate the management of soft ware construction, such as the SLOC (Source Lines of Code), [11] and FP (Function Points) [12] technique. These methods seek to evaluate comprehension according to features present in code lines or function points, making a proportional relation between them. In this paper, we highlight the approach proposed by [13] that pieces of evidence the complexity of the software linking it to the characteristics present in the cases of use of the system. Some factors are relevant to this approach: At first, the actors of the system are assigned graded labels, such as simple, medium and complex. This assessment lot depends on the synergy between business and systems analysis team to not measure undue influence on the actors who will be using the system. The next step is to set the Unweighted Weight of Actors (UAW) which is calculated by summing the weights of all actors [13]. The same procedure is performed with the use cases, characterizing them as simple, medium and complex, depending on the number of steps to run on your mainstream and alternate flows. In short, the fewer steps that are performed to complete an activity in a use case, the less complex it will be for the system and consequently to be developed. In the same way, as in the actors, the Unadjusted Use-Case Weight (UUCW) is calculated as the sum of the weights of all use cases. Then the UAW is added to the UUCW to produce the Unadjusted Use-Case Points (UUCP) [13]. Finally, the Evaluation Points are then adjusted according to the Technical Complexity Factors (TCF), such as the number of people involved, the client engagement, the level of knowledge of the team, among others, and environmental factors (EF) that external threats, such as crises, lack of corporate redemption, socio-economic or political events, are linked. Environmental factors and their weights were imported from Function Point theory, and technical complexity factors. Thus, UCP = UUCP * TCF * EF [13]. Figure 1 shows an applied example of any evaluation of UCP. Note that the weights and values will vary according to the nature of the problem being solved and the experts involved in this analysis. This means that within a UCP analysis, which may seem complicated for an inexperienced team in software that works with banking applications, it will be simpler for a team that has already developed systems for another financial institution. This allows the weight of some analysis parameters to be different in the UCP analysis of two companies that work with software development.

## 2.3. Software effort estimation

There are several comparative studies between the techniques of definition or estimation of efforts in software construction, as in [15]. Empirical studies using artificial neural networks were also applied in the prediction of CPU [16]. It should be noted that studies related to software testing efforts were also applied using UCP [17].

## 2.4. Intelligence models used in predicting use-case points

Intelligent models use the concepts of artificial intelligence to solve problems of various natures. These problems involve situations that business and human beings constantly experience. They can act in the pattern classification identifying images on web [18], license plate [19], saliency map estimation [20], image recognition [21, 22, 23] image recognition in medical domain [24], text recognition [25], fault identification [26], among others. Already in the data mining part of forecasting development recent work was developed by [27, 28, 29, 30].They use clever techniques to help managers and developers to understand the patterns involved in software development.Fuzzy rules-based fuzzy systems were used in [31] to perform the estimation of effort in the production of software using the UCP concepts. Already in the work of [32] the concepts of linear regression and perceptron were combined to solve the problems of estimation of efforts. Already in the works of [33] the focus was the same because the concepts of a cascade conceived the artificial neural network used. Finally, the model proposed by [34] worked to find the software effort with a hybrid two-layer model, where the first layer was composed of fuzzy neurons and the second layer composed of an artificial neuron.

| Factor | Weight | Value | Weight * Value |
|---|---|---|---|
| T1 Distributed system | 2 | 0 | 0 |
| T2 Response or throughput performance objectives | 2 | 0 | 0 |
| T3 End-user efficiency | 1 | 5 | 5 |
| T4 Complex internal processing | 1 | 1 | 1 |
| T5 Reusable code | 1 | 3 | 3 |
| T6 Easy to install | 0.5 | 4 | 2 |
| T7 Easy to use | 0.5 | 4 | 2 |
| T8 Portable | 2 | 1 | 2 |
| T9 Easy to change | 1 | 4 | 4 |
| T10 Concurrent | 1 | 3 | 3 |
| T11 Includes security features | 1 | 3 | 3 |
| T12 Provides access for third parties | 1 | 0 | 0 |
| T13 Special user training facilities are required | 1 | 5 | 5 |
| TFactor | | | 30 |

Figure 1: Example UCP [14]

It should be noted that in the model addressed by [31], fuzzy rules were also generated, but the training approach proposed in this article, as it was used with few dimensions, differs from the proposal of this work that intends to use all dimensions of the problem. The following are presented intelligent model architectures that were used to support this type of effort prediction, with emphasis on the models proposed in [31], and [32]

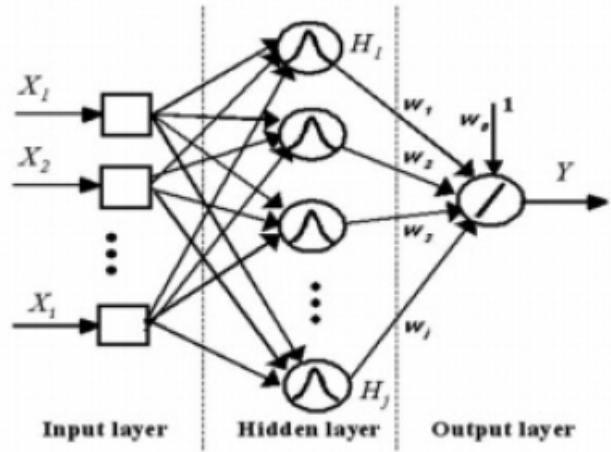

Figure 2: Effort model architecture [35]

## 2.5. Fuzzy Neural Network

Fuzzy neural networks are characterized by neural networks formed of fuzzy neurons [36].Thus a Fuzzy neural network can be defined as a fuzzy system that is trained by an algorithm provided by a neural network. The use of hybrid systems allows the joint use of advantageous properties that each one of the approaches can promote to solve the problem. Fuzzy systems are known for their ability to transform data into more coherent information for human understanding, especially if model responses will serve to construct expert systems. In the case of artificial neural networks, their ability to adapt to various types of training, learning techniques can perform activities commonly practiced by humans in a coherent way. Given this analogy, the union of the neural network with the fuzzy logic comes with the intention of softening the deficiency of each of these systems, making us have a more efficient, robust and easy to understand a system [37].

## 2.6. ANFIS

The Fuzzy Inference System (FIS) is a computational model based on the concepts of fuzzy set theory capable of generating fuzzy rules of the IF/THEN style and fuzzy reasoning. Its structure has three fundamental steps: a rule base, a database, and a reasoning base [6]. This type of model is capable of performing a non-linear mapping from its input space to the output space. This mapping is accompanied by many fuzzy rules IF/THEN, where each one describes the local behavior of the mapping. In these rules, the antecedent implements a multidimensional neural partition according to the type of algorithm chosen (grid, decision tree or grouping).

## 3. Fuzzy Neural Network software effort prediction using concepts of points of use cases.

### 3.1. Network architecture

The fuzzy neural network described in this section is composed of three layers, moreover, derives from the work of [38]. In the first layer, fuzzification is used through the concept of Anfis model. The membership functions adopted in the first layer are equally spaced and can be of the Gaussian or Triangular type. Already in the second layer the logical neurons of the and neuron type, different from the or neuron adopted in [5]. These neurons have weights and activation functions determined at random and through t-norms and s-norms to aggregate the neurons of the first layer. To define the weights that connect the second layer with the output layer, the concept of a extreme learning machine [8] is used to act on the neuron with a healthy activation function. And neuron is used to construct fuzzy neural networks in the second layer to solve pattern recognition problems and bring interpretability to the model. Figure 4 illustrates the feed forward topology of the fuzzy neural networks considered in this paper. The first layer is composed of neurons whose activation functions are membership functions of fuzzy sets defined for the input variables using Anfis. For each input variable $x_{ij}$, $L$ clouds are defined $A_{lj}$, $1 = 1$ L whose membership functions are the activation functions of the corresponding neurons. Thus, the outputs of the first layer are the membership degrees associated with the input values, i.e., $a_{jl} = \mu^A$ for $j = 1...$ $N$ and $l = 1$ $L$, where $N$ is the number of inputs and $L$ is the number of fuzzy sets for each input results by Anfis [5]. The second layer is composed by $L$ fuzzy and neurons. Each neuron performs a weighted aggregation of some of the first layer outputs. This aggregation is performed using the weights $w_{il}$ (for $i = 1$ $N$ and $l = 1$ $L$). For each input variable $j$, only one first layer output $a_{jl}$ is defined as input of the $l$-th neuron. So that **w** is sparse, each neuron of the second layer is associated with an input variable. Finally, the output layer is composed of one neuron whose activation functions are leaky ReLu [39]. The output of the model is:

$$y = \sum_{j=0}^{l} fleakyReLU(z_l v_l)$$

where $z_0 = 1$, $v_0$ is the bias, and $z_j$ and $v_j$, $j = 1, ..., l$ are the output of each fuzzy neuron of the second layer and their corresponding weight, respectively. Figure 4 presents an example of FNN architecture proposed in this paper.

This is an improved function of the ReLU function [40] because a small linear component is inserted at the input of the neuron. This type of change allows small changes to be noticed and neurons that would be relevant to the model are not discarded. Its function is expressed by [39]:

$$f_{LeakyReLU}(x, \alpha) = max(\alpha x, x) \qquad (2)$$

The logical neurons used in the second layer of the model are of the andneu- ron type, where the input signals are individually combined with the weights and performed the subsequent global aggregation. The andneuron used in this work can be expressed as [41]:

$$z = AND(w; z) = T_{i=1}^{n}(w_i s x_i) \qquad (3)$$

where $T$ are *t-norms*, $s$ is a *s-norms*. Fuzzy rules can be extracted from and- neurons according to the following example:

Rule$_1$: If $x_{i1}$ is $A^1$ with certainty $w_{11}$...
and $x_{i2}$ is $A^2$ with certainty $w_{21}$...

Then $y_1$ is $v_1$

$Rule_2$ : If $x_{i1}$ is $A^1$ with certainty $w_{12}$...

and $x_{i2}$ is $A^2$ with certainty $w_{22}$...

Then $y_2$ is $v_2$

$Rule_3$ : If $x_{il}$ is $A^1$ with certainty $w_{13}$...

Then $y_3$ is $v_3$

$Rule_4$ : If $x_{i2}$ is $A^2$ with certainty $w_{23}$...

Then $y_4$ is $v_4$ (4)

These rules allow the creation of a building base for expert systems [37].

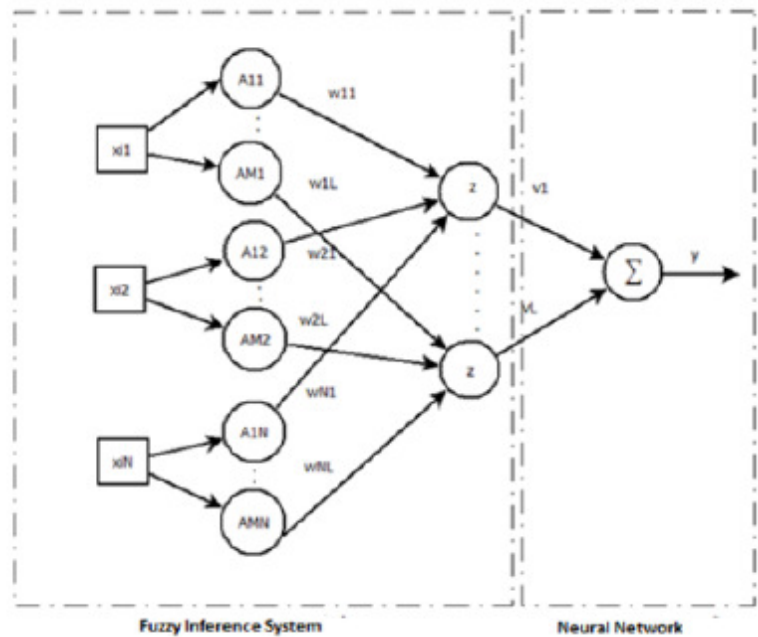

Figure 3: FNN architecture

## 3.2 Training Fuzzy Neural Network

The membership functions in the first layer of the FNN are adopted as Gaussian or Triangular. The number of neurons created with the input data partition is exponential between the number of membership functions and the number of features present in the problem database. The number of neurons $L$ in the first layer is defined according to the input data, and by the number of membership functions $M)$, defined parametrically. The second layer performs the aggregation of the $L$ neurons from the first layer through the and neurons.

After the construction of the $L$ and neurons the bolasso algorithm [42] is executed to select LARS (a regression algorithm for high-dimensional data that is proficient in measuring exactly the regression coefficients but also a subset of candidate regressors to be incorporated in the final model) using the most significant neurons (called $L_s$). The final network architecture is defined through a feature extraction technique based on $l_1$ regularization and resampling. The learning algorithm assumes that the output hidden layer composed of the candidate neurons can be written as [43]:

$$f(x_i) = \sum_{i=0}^{L_s} v_i z_i(x_i) = z(x_i)v \qquad (5)$$

where $\mathbf{v} = [v_0, v_1, v_2, ..., v_L]$ is the weight vector of the output layer and $\mathbf{z}$ $(xi) = [z_0, z_1(x_i), z_2(x_i)z_L(x_i)]$ the output vector of the second layer, for $z_0 = 1$.

In this context, $\mathbf{z}$ $(x_i)$ is considered as the non-linear mapping of the input space for a space of fuzzy characteristics of dimension $L_\rho$ [43].

The LARS algorithm can be used to perform the model selection since for a given value of $\lambda$ only a fraction (or none) of the regressors have corresponding nonzero weights. If $\lambda = 0$, the problem becomes unrestricted regression, and all weights are nonzero. As $\lambda_{max}$ increases from 0 to a given value $\lambda_{max}$, the number of nonzero weights decreases to zero. For the problem considered in this paper, the $z_l$ regressors are the outputs of the significant neurons. [42]. Bolasso procedure is summarized in Algorithm 1.

---

**Algorithm 1:** Bolasso- bootstrap-enhanced least absolute shrinkage operator

(b1) Let $n$ be the number of examples, (lines) in X: (b2) Show $n$ examples of (X, Y), uniformly and with substitution, called here (*Xsamp*, *ysamp*). (b3) Determine which weights are nonzero given a $\lambda$ value. (b4) Repeat steps b1: b3 for a specified number of bootstraps $b$. (b5) Take the intersection of the non-zero weights indexes of all bootstrap replications. Select the resulting variables. (b6) Revise using the variables selected via non-regularized least squares regression (if requested). (b7) Repeat the procedure for each value of $b$ bootstraps and $\lambda$ (actually done more efficiently by collecting interim results). (b8) Determine "optimal" values for $\lambda$ and $b$.

---

Subsequently, following the determination of the network topology, the pre- dictions of the evaluation of the vector of weights' output layer are performed. In this paper, this vector is considered by the Moore-Penrose pseudo Inverse [43]:

$$v = Z^+ y \qquad (6)$$

$Z^+$ is the Moore-Penrose pseudo Inverse of $z$, which is the minimum norm of the least squares solution for the output weights. synthesized as demonstrated in Algorithm 2. It has three parameters:

1- the number of grid size, $\rho$;
2- the number of bootstrap replications, $bt$;
3- 3- the consensus threshold, $\lambda$.

# 4. REGRESSION MODELS OF USE CASE POINT PROBLEMS

## 4.1 Assumptions and Initial Test Configurations

The tests performed in this paper seek to find a predictor model for the definition of the effort in hours for the construction of software using the fuzzy

---

Algorithm 2: Forecasting Effort -FNN training

1) Define the number os membership functions, $M$.
2) Define bootstrap replications, $bt$.
3) Define consensus threshold, $\lambda$
4) Calculate L neurons in the first layer using Anfis.
5) Construct L fuzzy neurons with Gaussian ou Triangular membership functions constructed with center and sigma values derived from Anfis.
6) Define the weights and bias of the fuzzy neurons randomly.
7) Construct L and neurons with random weights and bias on the second layer of the network by welding the L fuzzy neurons of the first layer.
8) **For all** K entries do
(8.1) Calculate the mapping $z_k(x_k)$ using and neurons
9) Select significant $L_s$ using the lasso bootstrap according to the settings of $bt$ and $\lambda$.
10) Estimate the weights of the output layer (6)
11) Calculate output y using leaky ReLU. (2)

---

neural network. The accuracy of the training and the test of the model will be realized by checking the values obtained by the model, comparing them with the expected result. In this context, they are evaluated through the mean square error (MSQE). The formula for defining your calculation is shown below:

$$RMSE = \frac{1}{N}(\sum_{k=0}^{n} y^k - y'^k)^{\frac{1}{2}} \qquad (7)$$

To perform the training, 30 repetitions were performed with the samples made available through the software construction effort study. The percentage is defined as 70 % of the samples allocated for training and the remaining 30 % for the test phase of the model. To avoid trends in the characteristics of each of the examples, a proposal was made where all the samples destined to the training and testing of the fuzzy neural network were randomly sampled. This ensures that there will be no dependencies of the data stream for the model results. All samples involved in the test were normalized with mean zero and variance 1. The activation functions of the third layer neuron are of the leaky ReLU type. The values of the bootstrap replicate, the decision consensus for the use of the bolus are, respectively, 16 and 0.7. These values were found to be optimal values using a 5 k-fold technique in preliminary tests. For fuzzy neural networks using equally spaced Gaussian or triangular functions, it was defined as 2 by the same k-fold process that the configuration parameters of the bolasso method were found

## 4.2. Database used in the tests.

Data set was collected by [4] from three software companies that provided the data (D1, D2, and D3) and is based on the following problem areas: Insurance, Government, Banks, and other domains. This database provides data such as the methodology used in software production, the complexity of weights for cases of uses and actors, and other metrics relevant to the evaluation of efforts. To be applied in the fuzzy neural network, the columns referring to the textual information were transformed into numerical values.

The Principal dimensions used in this evaluation are: donator, simple actor methodology, average actors, complex actors, actor weight, simple use case, medium use case, complicated use case, use case weight, technical complexity, the real effort 20 hours. The methodology was defined as 0 for traditional methods and 1 for agile methodologies.

## 4.3. Prediction tests on efforts in software construction.

Table I presents the results of the model proposed in this paper

Table 1: Accuracies of the FNN

| M type | L | $L_s$ | RMSE Train. | RMSE Test |
|---|---|---|---|---|
| Gaussian | 100.00 (0) | 27.83 (18.62) | 56.85 (28.42) | 205.13 (75.55) |
| Triangular | 100.00 (0) | 20.77 (12.90) | 58.28 (2,29) | 145.18 (45.15) |

After the tests, it was verified that the model behaves very efficiently in the prediction of efforts related to the construction of software. In the comparison between the two models, the one that uses triangular functions had the better result. The approach utilized all the dimensions provided in the database and had close results to the regression models proposed by [4] which used a subgroup of features of that database. This approach worked on the final network with an adequate number of neurons.

The following is the results of the prediction performed by the fuzzy neural network. This example demonstrates through graphs, relevant characteristics of the answers obtained by the model In figure 4 results of training. Fig 5 the results of tests.

## 4.4 Interpretability of the problem based on fuzzy rules.

Figure 6 below shows the representation of Anfis used in the first layer of the model. The following are examples of rules with the bases used. Each mem- bership function can receive values to be consistent with the analyzed context.

If (Donator is complex) and (Methodology is Agile) and (SimpleActor is large) and (MediumActor is small) and (Complexactor is small) and (Weigh tActor is high) and (usecaseSimple is large)and (usecaseMedium is small) and (usecaseComplex is large) and -0.2482 then (effort is 1762.18) (8) (Weight Use Case is high) and (Techinical complexity is low) and (complexity factors is low) and (realeffort20hours is high) with certainty -0.0307 then (effort is 167.863) 8. If (Donator is small) and (Methodology is traditional) and (Simple Actor is large) and (Medium Actor is small) and (Complex actor is small) and (Weight Actor is low) and (use case Simple is large) and (use case Medium is small) and (use case Complex is small) and (Weight Use Case is high) and (Techinical complexity is high) and (complexity factors is high) and (realeffort20hours is low) with certainty

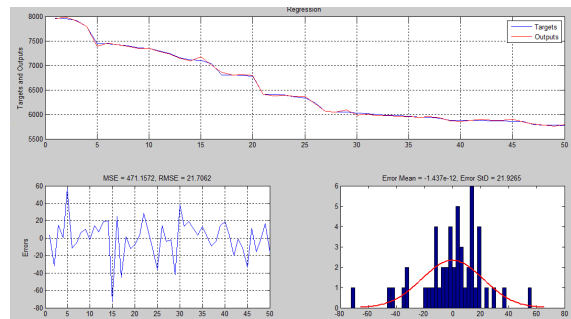

Figure 4: Predict Train Results example

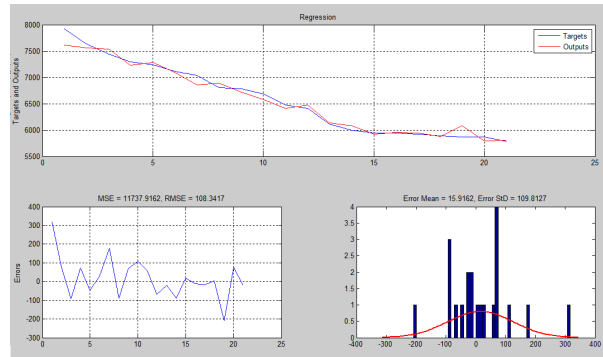

Figure 5: Predict Test Results

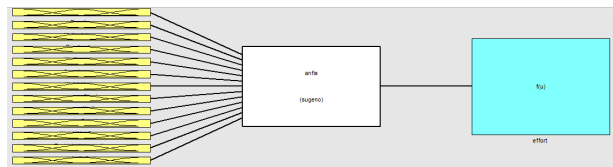

Figure 6: Anfis Results

## 5. CONCLUSION

The results presented by the fuzzy neural network model are promising for the prediction of effort in software construction. In projects with more than 5000 service hours, averaging around 150 hours may be a fundamental way of helping managers predict. The advantage proposed by this work was to use all the arguments present in the UCP since as new projects are analyzed, the data of new evaluations can change the influence of the weights and correlations involved in the variables made available for the test. This type of generated fuzzy rules can be more adapted than works that made a selection of characteristics, especially if the nature of this discarded information becomes relevant to the problem. The fuzzy neural network approach allows knowledge to be extracted from the database to foster training and knowledge for software development firms. With traditional approaches, we obtain the estimation of software effort, but the results are not interpretable because they are a black box problem.

It can be verified that the construction of fuzzy rules allows managers, em- ployees, and developers to find the logical relationships between the data set and the problem to be solved.

With the fuzzy rules found, training, hiring, and workforces can be allocated in important contexts to avoid time for software development to be high. Future work may be performed on other models of fuzzy neural networks, which use other neurons or clustering methods. Other activation functions can be applied to verify the accuracy of the model.

### ACKNOWLEDGMENT

The thanks of this work are destined to CEFET-MG and UNA.**Conflict of Interest:** All authors declare no conflict of interest.